\documentclass[conference, letterpaper]{ieeeconf}
\IEEEoverridecommandlockouts
\usepackage{cite}
\usepackage{amsmath,amssymb,amsfonts}
\usepackage{graphicx}
\usepackage{textcomp}
\usepackage{xcolor}
\def\BibTeX{{\rm B\kern-.05em{\sc i\kern-.025em b}\kern-.08em
    T\kern-.1667em\lower.7ex\hbox{E}\kern-.125emX}}

\let\labelindent\relax  
\usepackage{algorithm}
\usepackage{algorithmicx}  
\usepackage[noend]{algpseudocode}
\usepackage{amsfonts}
\usepackage{amsmath}
\usepackage{amsthm}
\usepackage{bm}
\usepackage{booktabs}
\usepackage{diagbox}
\usepackage{dsfont} 
\usepackage{enumitem}
\usepackage{etoolbox}
\usepackage{graphicx}  
\usepackage{hyperref}  
\usepackage{lipsum}  
\usepackage{mathtools}
\usepackage{multirow}
\usepackage{nicefrac}
\usepackage[final]{pdfpages}
\usepackage{siunitx}
\usepackage[caption=false]{subfig}
\usepackage{tabu}
\usepackage{threeparttable}  
\usepackage{tikz}
\usepackage{xcolor}
\usepackage{xfrac}

\usepackage{pgfplots}   
\pgfplotsset{compat=newest}
\usepgfplotslibrary{groupplots}
\usepgfplotslibrary{fillbetween}
\usetikzlibrary{pgfplots.statistics} 

\graphicspath{{img/}}
\hypersetup{colorlinks=false}

\newcommand{\method}[1]{\textsc{#1}}

\newcommand{\rom}[1]{\MakeUppercase{\romannumeral #1}}

\definecolor{githubColor}{HTML}{2EA44F}

\definecolor{newGray}{HTML}{808080}

\definecolor{backGray}{rgb}{0.3, 0.3, 0.3}%
\definecolor{matlabYellow}{rgb}{0.9290, 0.6940, 0.1250}%
\definecolor{matlabPurple}{rgb}{0.4940, 0.1840, 0.5560}%
\definecolor{matlabLBlue}{rgb}{0.3010, 0.7450, 0.9330}%
\definecolor{matlabGreen}{rgb}{0.4660, 0.6740, 0.1880}%
\definecolor{matlabRed}{rgb}{0.8500, 0.3250, 0.0980}%
\definecolor{matlabBlue}{rgb}{0, 0.4470, 0.7410}%
\definecolor{matlabDarkRed}{rgb}{0.6350 0.0780 0.1840}%

\definecolor{colorCircle}{HTML}{0072BD}

\definecolor{colorRect}{HTML}{D95319}

\newcolumntype{O}[1]{S[detect-weight, mode=text, table-format=#1]}

\renewcommand{\bfseries}{\fontseries{b}\selectfont} 
\robustify\bfseries             
\newrobustcmd{\B}{\bfseries} 

\makeatletter
\AddToHook{env/tabular/begin}{\let\input\@@input}
\makeatother

\newcommand\copyrighttext{\footnotesize \textcopyright~2024 IEEE. Personal use of this material is permitted. Permission from IEEE must be obtained for all other uses, in any current or future media, including reprinting/republishing this material for advertising or promotional purposes, creating new collective works, for resale or redistribution to servers or lists, or reuse of any copyrighted component of this work in other works.
DOI: \href{https://doi.org/10.1109/IV55156.2024.10588448}{10.1109/IV55156.2024.10588448}
}

\newcommand\copyrightnotice{%
    \begin{tikzpicture}[remember picture,overlay]%
 	\node[anchor=south, xshift=-0pt, yshift=20pt] at (current page.south)%
 	{\fbox{\parbox{\dimexpr\textwidth-\fboxsep-\fboxrule\relax}{\copyrighttext}}};%
 	\end{tikzpicture}%
}

\newtheoremstyle{tstyle}
  {}
  {}
  {\itshape}
  {}
  {\bfseries}
  {.}
  { }
  {\thmname{#1}\thmnumber{ #2}\thmnote{ (#3)}}%
\theoremstyle{tstyle}

\newcommand{\mbeq}{\overset{!}{=}}

\newcommand{\mat}[1]{\boldsymbol{#1}}

\renewcommand{\vec}[1]{\boldsymbol{#1}}

\newcommand{\inv}{^\text{\rmfamily \textup{-1}}}

\newcommand{\trans}{^\text{\rmfamily \textup{T}}}

\newcommand{\sol}[1]{\hat{#1}}

\newcommand{\pspace}{\,}  

\title{Globally Optimal GNSS Multi-Antenna Lever Arm Calibration}
\author{Thomas Wodtko and Michael Buchholz%
\thanks{This work was financially supported by the State Ministry of Economic Affairs Baden-Württemberg (project U-Shift\,II, AZ\,3-433.62-DLR/60).}%
\thanks{Both authors are with the Institute of Measurement, Control and Microtechnology, Ulm University, Albert-Einstein-Allee 41, 89081 Ulm, Germany {\tt\footnotesize \{firstname\}.\{lastname\}@uni-ulm.de}}%
}

\begin{document}

\maketitle

\begin{abstract}
Sensor calibration is crucial for autonomous driving, providing the basis for accurate localization and consistent data fusion.
Enabling the use of high-accuracy GNSS sensors, this work focuses on the antenna lever arm calibration. 
We propose a globally optimal multi-antenna lever arm calibration approach based on motion measurements.
For this, we derive an optimization method that further allows the integration of a-priori knowledge. 
Globally optimal solutions are obtained by leveraging the Lagrangian dual problem and a primal recovery strategy.
Generally, motion-based calibration for autonomous vehicles is known to be difficult due to cars' predominantly planar motion.
Therefore, we first describe the motion requirements for a unique solution and then propose a planar motion extension to overcome this issue and enable a calibration based on the restricted motion of autonomous vehicles. 
Last we present and discuss the results of our thorough evaluation. 
Using simulated and augmented real-world data, we achieve accurate calibration results and fast run times that allow online deployment.
\end{abstract}

\section{Introduction}
\copyrightnotice
With increasing autonomy, robotic systems are equipped with more and more sensors.
Multiple sensors of different types are used to, e.g., cover a wide area or allow redundant coverage.
A perception pipeline incorporates all sensor data using suitable processing approaches.
However, before sensor data is used, all sensors must be extrinsically calibrated to allow accurate data fusion.
Recent concepts of autonomous vehicles~\cite{ushift2022} further require that such calibration data is at least verified or updated online since additional sensors are added during run-time.

This work focuses on a specific combination of sensor types: Inertial Measurement Units (IMUs) and GNSS receivers~\cite{hossein2018deep}.
Such a sensor combination relies on accurately specifying the relative position between each GNSS antenna and the IMU origin, called the antenna lever arm.
Using multiple GNSS sensors allows one to e.g. directly measure an autonomous vehicle's orientation.
Given the disjunct data types of both sensors, this information cannot be determined directly from a single measurement~\cite{hong2004car, stovner2019gnss}.
An often-used approach is to measure the distance using a laser rangefinder; however, this requires an additional external device, and the procedure can become cumbersome when the sensor housing is not easily accessible, e.g., when the IMU is placed inside the trunk. 
Further, the exact phase center of an antenna is not always known to the user.
\begin{figure}[t]
    \centering
    \input{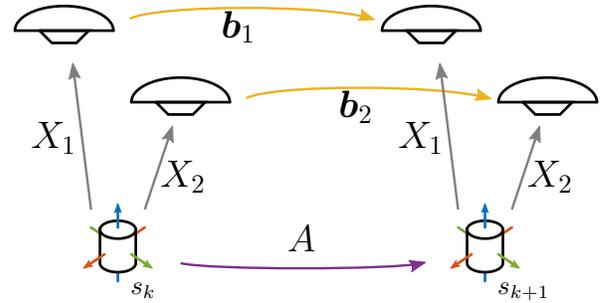}
    \caption{%
        A calibration scenario for a single time step is illustrated.
        An IMU sensor, together with two antennas, is moving from station $s_k$ (left) to $s_{k+1}$ (right).
        The calibration goal is to obtain the transformation $X_i$ for each antenna.
        The motion of the IMU sensor (cylinder) is represented by $A$ (purple arrow).
        $X_1$ and $X_2$ (gray arrows) are the transformations from the IMU sensor to the respective antenna.
        $\vec{b}_{1}$ and $\vec{b}_{2}$ (yellow arrows) represent the motion of the respective antenna.
        As a GNSS sensor does not measure orientation, only translational motions are available.
    }
    \label{fig:gala}
    \vspace{-0.5cm}
\end{figure}
The task of calibrating an antenna lever arm was addressed in previous works.
Some approaches derive a motion model for the robotic system at hand and apply a Kalman filter to estimate the antenna lever arm offset~\cite{hong2004car, stovner2019gnss}.
Others use neural networks to solve the calibration~\cite{montalbano2018comparison}.

Another class of calibration based on motion measurements is the so-called hand-eye calibration.
As shown in our previous work~\cite{horn2021online, wodtko2021globally, horn2023extrinsic}, this type of calibration is well-suited for the calibration of different sensor types due to the use of motion transformations.
For example, motion data can be obtained from data of different sensor types using open source software~\cite{murartal2015orbslam}.
However, a GNSS sensor only measures the position, not the orientation of an antenna; thus, 6D motion is not available, and hand-eye calibration cannot directly solve the antenna lever arm calibration task.
This work aims to derive a Quadratically Constrained Quadratic Programming (QCQP) formulation, similar to the hand-eye formulation of~\cite{horn2021online}, for the antenna lever arm calibration focusing on autonomous vehicles.
For this, motion requirements for a unique solution are derived, which aim to complement previous investigations by the authors of~\cite{hong2004car, hong2005observ}.
The calibration scenario is illustrated in Fig.~\ref{fig:gala}, and similarities to the hand-eye calibration (see ~\cite{wodtko2021globally}) can be made out.
Summarizing our work in this paper, we propose
\begin{itemize}[topsep=0pt]
    \item a novel Quadratically Constrained Quadratic Programming (QCQP) formulation for the antenna lever arm calibration,
    \item a definition of motion requirements, and
    \item a respective extension of the Python-based open-source calibration library Excalibur~\cite{excalibur}.
\end{itemize}

\section{Related Work}
\label{sec:relatedWork}

Most available approaches for the antenna lever arm calibration define a motion model and use a Kalman filter to estimate the antenna offset~\cite{hong2004car, hong2005observ, stovner2019gnss, montalbano2018comparison}.
For this, the motion of the robotic system at hand must be described accurately to allow the estimation of the antenna offset using an error model.
While such approaches can incorporate measurement uncertainties, they must be adjusted to every specific target platform. 
For example, any motion restriction must be considered in the respective model.

Generally, neural networks can also solve the antenna lever arm calibration.
The authors of~\cite{montalbano2018comparison} demonstrated the applicability, but they also state that their results should be "viewed as a lower bound on performance."
More data and evaluations are required for an extensive and fair comparison to classical approaches. 
Nevertheless, the results in~\cite{montalbano2018comparison} indicate that Kalman filter-based approaches outperform those implying neural networks.

This work derives a different formulation for the antenna lever arm calibration.
Based on hand-eye calibration~\cite{giamou2019certifiably, horn2021online}, the formulation decouples the motion estimation and the antenna lever arm calibration.
Hence, the underlying motion model does not need to consider motion restrictions of, e.g., autonomous vehicles.
Instead, it can be more sophisticated and cover multiple motion patterns of different platforms without the need to adjust the calibration procedure.
Further, as shown in~\cite{horn2021online}, the globality of a solution can be verified and sometimes even guaranteed.
Additionally, prior knowledge can be integrated into the optimization.
For example, motion restrictions of planar motion that lead to an unobservable mode can be compensated for, similar to~\cite{horn2021online, horn2023extrinsic}. 
To the best of our knowledge, no similar approach for the antenna lever arm calibration has been proposed before.

The authors of~\cite{hong2004car, hong2005observ, lee2005observ} investigated the observability of the lever arm.
They examined the observability analytically, with simulations, and in real-world experiments to assess different motion patterns concerning the final estimation error.
However, the statement on what specific motion is required is somewhat vague.
Instead, recommendations describing what kind of motion benefits the calibration are given.
The authors of~\cite{lee2005observ} claim that only accelerated motion improved the lever arm estimation and that motion of constant angular velocity had no influence on estimation in their experiments.
In contrast, this work will show that an accurate antenna lever arm calibration is possible without considering acceleration at all; 
and instead, if adequately excited, the sole consideration of the orientation changes of the IMU sensor over time is sufficient.
More specifically, this work analyses the motion requirements from a different perspective and gives a precise formulation of the required motion.
Given these requirements for the approach proposed in this work, we aim to extend the statements of~\cite{lee2005observ,hong2005observ} to allow a user to ensure sufficient excitement when performing a calibration.
\section{Foundations}

This work uses homogeneous matrices (HMs) to derive an antenna lever arm calibration formulation.

Generally, transformations are defined frame-forward and referred to as functions denoted by uppercase italic letters $T$.
The composition of two transformations is denoted by $T_a \circ T_b$.
Vectors and matrices are denoted by lowercase and uppercase bold letters, respectively, e.g. $\vec{x}$ and $\mat{X}$.

\subsection{Homogeneous Matrices}

This section outlines the structure and basics of HMs.
For a more detailed description, we refer to~\cite{mccarthy1990intro}.
A HM $\mat{T} \in \text{SE}(3)$ representing the rigid transformation $T$ in 3D space, given by
\begin{align}
\label{eq:HMs}
\mat{T} = 
\begin{bmatrix}
\mat{R} & \vec{t} \\
\mat{0} & 1
\end{bmatrix} \pspace ,
\end{align}
consists of a rotation matrix $\mat{R} \in \text{SO}(3)$ and an translation vector $\vec{t} \in \mathbb{R}^3$.
The transformation of a point $\widetilde{\vec{p}} = T(\vec{p}) \pspace , \pspace \vec{p} \in \mathbb{R}^3$, or a rotation axis $\widetilde{\vec{a}} = T(\vec{a}) \pspace , \pspace \vec{a} \in \mathbb{R}^3$, is applied by multiplying the respective homogeneous vector to the right side of the HM, denoted by
\begin{align}
\label{eq:hmTransform}
    \begin{bmatrix}
        \widetilde{\vec{p}} & 1
    \end{bmatrix}\trans
    = \mat{T} \pspace 
    \begin{bmatrix}
        \vec{p} & 1
    \end{bmatrix}\trans \quad \text{and}
    \quad
    \begin{bmatrix}
        \widetilde{\vec{a}} & 0
    \end{bmatrix}\trans
    = \mat{T} \pspace 
    \begin{bmatrix}
        \vec{a} & 0
    \end{bmatrix}\trans \pspace ,
\end{align}
respectively.
At this point, it shall be mentioned that a frame-forward transformation $T_{ab}$ from frame $a$ to frame $b$ transforms a frame from $a$ to $b$. 
However, a point is transformed in the opposite direction from $b$ to $a$.
By this, the composition of two transformations $T_{ac} = T_{ab} \circ T_{bc}$ is represented by the multiplication of the respective HMs $\mat{T}_{ac} = \mat{T}_{ab} \mat{T}_{bc}$.
A point-forward definition would require a change of the multiplication order.
For simplicity reasons, if a transformation $B$ does not contain any rotation, i.e., $\mat{R}_B = \mat{I}$, the translation is denoted by $\vec{b}$ instead of $\vec{t}_{B}$.

\subsection{Quadratically Constrained Quadratic Programming}

Here, the fundamentals of QCQP required by the optimization in Section~\ref{sec:method} are explained.
Given a positive semi-definite matrix $\mat{Q} \in \mathbb{R}^{n \times n}$, a purely quadratic cost or objective function $J: \mathbb{R}^n \rightarrow \mathbb{R} \pspace ; \pspace J(\vec{z}) := \vec{z}\trans \mat{Q} \vec{z}$ can be constructed.
Further, an equality constraint function $\vec{g}: \mathbb{R}^n \rightarrow \mathbb{R}^m$ containing $m$ constraints, which must be satisfied by the objective variable $\vec{z}$, can be defined.
The QCQP optimization problem for the objective variable $\vec{z} \in \mathbb{R}^n$ is then defined by
\begin{subequations}
\label{eq:primalProb}
\begin{alignat}{2}
    & \!\min_{\vec{z}} & ~ & J(\vec{z})\\
    & \text{w.r.t.}	 & ~ & \vec{g}(\vec{z}) \mbeq \vec{0} \pspace .
\end{alignat}
\end{subequations}
In later sections, two types of constraints are important: purely quadratic and homogenization constraints.
They are defined by
\begin{align}
\label{eq:quadAndHomConst}
    g_q(\vec{z}) = \vec{z}\trans \mat{P}_q \vec{z} \quad \text{and} \quad g_h(\vec{z}) = 1 + \vec{z}\trans \mat{P}_h \vec{z} \pspace ,
\end{align}
respectively, with the matrices $\mat{P}_q, \mat{P}_h \in \mathbb{R}^{n \times n}$.
While $\mat{P}_q$ is task-specific, the matrix $\mat{P}_h$ depends only on the homogenization index of the objective variable. 
If it is the last index in $\vec{z}$, all entries of $\mat{P}_h$ are zero, except $p_{h,(n,n)} = -1$.
As required later, $m_q$ quadratic and one homogenization constraints can be combined to a single constraint function $\vec{g}(\vec{z}) = \begin{bmatrix} g_{h}(\vec{z}), g_{q,1}(\vec{z}) \dots g_{q,m_q}(\vec{z})\end{bmatrix}\trans$.

Due to non-linear constraints, the resulting problem is not convex; thus, finding a globally optimal solution is not trivial.
Therefore, the derivation of the Lagrangian dual problem is summarized next, analogously to~\cite{horn2021online,wodtko2021globally}.
Given the Lagrangian function and $\vec{\lambda} := \begin{bmatrix} \lambda_1 & \dots & \lambda_m \end{bmatrix}\trans \in \mathbb{R}^{m}$
\begin{align}
    L(\vec{z},\vec{\lambda}) = \vec{z}\trans\mat{Z}(\vec{\lambda})\vec{z} + \lambda_1  \pspace ,
\end{align}
with $\mat{Z}(\vec{\lambda}) := \mat{Q} + \lambda_1 \mat{P}_{h} + \sum_{i=1}^{k} \lambda_i \mat{P}_{q,i}$, the dual problem is
\begin{subequations}
\label{eq:dualProb}
\begin{alignat}{2}
&\!\max_{\vec{\lambda}}  &\quad& \lambda_1\\
&\text{w.r.t.} &      & \mat{Z}(\vec{\lambda}) \succeq \mat{0}  \pspace .
\end{alignat}
\end{subequations}
From dual theory~\cite{boyd2004convex}, it is known that the Lagrangian dual problem is always convex; hence, a guaranteed globally optimal solution $\sol{\vec{\lambda}}$ can be obtained using available solvers like~\cite{diamond2016cvxpy}.
$\sol{\lambda_1}$ is a lower bound for the primal optimal value $J(\sol{\vec{z}})$.
The difference between the primal and dual optimal values is called the duality gap.
If the duality gap is zero, we know that the primal solution $\sol{\vec{z}}$ is globally optimal.
The following section describes how a primal solution can be recovered from a dual solution.

\subsection{Recover Primal from Dual Solution}

As described in~\cite{horn2021online}, a certifiably globally optimal primal solution $\sol{\vec{z}}$ can be recovered from the null space of $\mat{Z}(\sol{\vec{\lambda}})$.
If the null space has dimension $1$, the solution is recovered by normalizing the null space vector using the homogenization constraint of \eqref{eq:quadAndHomConst}.
If the null space has dimension $2$, the solution is recovered using the equations derived in \cite{daniilidis1999handeye}.
In both cases, the resulting primal solution is guaranteed to be globally optimal.
If the null space dimension is greater than two, the globality of a solution can only be verified.
For this, the non-convex primal problem is optimized directly, with the normalized null space vector corresponding to the largest eigenvalue as the initial solution.
After recovering the primal solution, the duality gap must be checked.
Only if the duality gap is zero, it is verified that the recovered solution is globally optimal.

\section{Method}
\label{sec:method}

In this section, the GNSS multi-antenna lever arm calibration problem is first defined in detail.
Then, the proposed calibration procedure is derived for single and multiple antennas.
Afterward, the extensions of~\cite{horn2021online} and~\cite{horn2023extrinsic} are picked up and fitted to the calibration of this work.
Last, the optimization problem is analyzed to formulate precise motion requirements for a unique calibration solution.

\subsection{Problem Formulation}
\label{sec:problemFormulation}

The GNSS multi-antenna lever arm calibration aims to obtain each antenna's position relative to the IMU sensor frame.
An exemplary scenario at a single time step is illustrated in Fig.~\ref{fig:gala}.
For each step $k \in \mathbb{N}_0$, an IMU measurement $A_{k} \in \text{SE}(3)$ and GNSS measurements $\vec{b}_{i,k} \in \mathbb{R}^3$ for each antenna $i$ are given.
The orientation of the GNSS frame is assumed to equal the IMU frame orientation (see Section~\ref{sec:motionRequirements} for details).
Similar to the hand-eye problem, see~\cite{horn2021online}, a cost function is derived based on transformation cycles.
However, no 6D motion is available for the GNSS sensor.
Instead, the cycle is defined by
\begin{align}
\label{eq:trafoCycle}
    X_{i}(\vec{b}_{i,k}) = (A_{k} \circ X_{i})(\vec{0}) \pspace .
\end{align}
Here, the position of an antenna after moving is projected to the IMU frame before motion using both available paths.
In this work, it is assumed that IMU measurements are available as motion transformations.
However, in general, IMU sensors do not measure position or motion transformations directly but rather inertial entities, like rotational speed and translational acceleration.
Thus, to obtain transformations, a preprocessing step might be required.

\subsection{Single-Antenna Calibration}
\label{sec:singleAntenna}
Using the Eqs.~\eqref{eq:HMs}, and~\eqref{eq:hmTransform}, Eq.~\eqref{eq:trafoCycle} for a single antenna at a single time step can be formed into
\begin{subequations}
\begin{alignat}{2}
\begin{bmatrix}
    \mat{I} & \vec{x} \\
    \vec{0} & 1
\end{bmatrix}
\begin{bmatrix}
    \vec{b} \\
    1
\end{bmatrix}
 & = 
\begin{bmatrix}
\mat{R}_{A} & \vec{t}_{A} \\
\vec{0} & 1
\end{bmatrix}
\begin{bmatrix}
    \mat{I} & \vec{x} \\
    \vec{0} & 1
\end{bmatrix}
\begin{bmatrix}
    \vec{0} \\
    1
\end{bmatrix} \pspace , \\[8pt]
\vec{b} + \vec{x} & = \mat{R}_{A} \vec{x} + \vec{t}_{A} \pspace .
\end{alignat}
\end{subequations}
Using a homogenization variable $\mu \in \mathbb{R}$ with $\mu^2 = 1$, the equation can be rearranged to
\begin{align}
\label{eq:singleCycle}
\mat{M} \pspace \vec{z} := 
    \begin{bmatrix}
        \mat{R}_{A} - \mat{I} & \vec{t}_{A} - \vec{b}
    \end{bmatrix}
    \begin{bmatrix}
        \vec{x} \\
        \mu
    \end{bmatrix}
    = \vec{0} \pspace .
\end{align}
Now, the matrix $\mat{M}_t \in \mathbb{R}^{3 \times 4}$ contains all motion measurements of the current time step. 
The vector $\vec{z} \in \mathbb{R}^{4}$ contains the desired calibration result and, at the last index, the homogenization variable.
Given $l$ time steps, a respective quadratic cost function $J: \mathbb{R}^4 \rightarrow \mathbb{R} \pspace ; J(\vec{z}) := \vec{z}\trans \mat{Q} \vec{z}$ is constructed with $\mat{Q} \in \mathbb{R}^{4 \times 4}$ defined by
\begin{align}
    \mat{Q} := \sum_{k=1}^{l} \mat{M}_{k}\trans \mat{M}_{k} \pspace .
\end{align}
Using this cost function without any purely quadratic constraint ($m_q = 0$), the solution to the optimization problem~\eqref{eq:primalProb} obtains the result of a GNSS singe-antenna lever arm calibration.

\subsection{Multi-Antenna Calibration}
\label{sec:multiAntennacalibration}
\begin{figure*}
    \captionsetup[subfloat]{labelfont=normalsize,textfont=normalsize}
    \subfloat[Structure of $\sum\limits_{i=0}^{o} \mat{M}_{k}^{i}$]{%
        \label{fig:matricStructure:first}%
        \def\colorPalleteRa{{32,62,74,57,28,46,45,56,21}}
\newcommand\ra{} 
\def\ra[#1](#2,#3)(#4,#5){%
    \foreach \x [count=\n from 0] in {0,1,2}
    {%
        \foreach \y [evaluate=\myMix using ({\colorPalleteRa[\n*3 + \y]}+20)]in {0, 1, 2} {%
            \draw [#1, draw=none, fill=matlabGreen!\myMix!black] (#2+#4*\x, #3-#5*\y) rectangle ++(#4,-#5);
        }
    }
}

\def\colorPalleteT{{90,60,77}}
\newcommand\tvec{} 
\def\tvec[#1](#2,#3)(#4,#5)(#6,#7){%
    \foreach \step [evaluate=\myMix using ({\colorPalleteT[\step]})]in {0,1,2} {%
        \draw [#1, draw=none, fill=#4!\myMix!black] (#2+#6*\step*#5, #3+#7*\step*#5) rectangle ++(#5,-#5);
    }
}

\resizebox{0.43\linewidth}{!}{%
\begin{tikzpicture}[baseline=(current bounding box.center)]
\useasboundingbox (0,0.5) rectangle (14,-9.5);

    \node [scale=2] (rai) at (11+2.7/2, -0) {$\mat{R}_A - \mat{I}$};
    \ra[](11,-0.5)(0.9,0.9);

    \node [scale=2] (tb1) at (11+2.7/2, -4) {$\vec{t}_{A} - \vec{b}_{1}$};
    \tvec[](11,-4.5)(yellow, 0.9)(1,0);
    \node [scale=2] (tb1) at (11+2.7/2, -6) {$\vec{t}_{A} - \vec{b}_{2}$};
    \tvec[](11,-6.5)(matlabDarkRed, 0.9)(1,0);
    \node [scale=2] (tb1) at (11+2.7/2, -8) {$\vec{t}_{A} - \vec{b}_{3}$};
    \tvec[](11,-8.5)(matlabBlue, 0.9)(1,0);

    \ra[](0,0)(1,1);
    \ra[](3,-3)(1,1);
    \ra[](6,-6)(1,1);
    \tvec[](9,0)(yellow, 1.0)(0,-1);
    \tvec[](9,-3)(matlabDarkRed, 1.0)(0,-1);
    \tvec[](9,-6)(matlabBlue, 1.0)(0,-1);
    
    \draw[draw=black] (0,0) rectangle ++(10,-9);
    \draw[draw=black] (3,0) - ++(0,-9);
    \draw[draw=black] (6,0) - ++(0,-9);
    \draw[draw=black] (9,0) - ++(0,-9);
    \draw[draw=black] (0,-3) - ++(10,0);
    \draw[draw=black] (0,-6) - ++(10,0);

    \node[scale=1.5] () at (1.5, 0.5) {1:3};
    \node[scale=1.5] () at (4.5, 0.5) {4:6};
    \node[scale=1.5] () at (7.5, 0.5) {7:9};
    \node[scale=1.5] () at (9.5, 0.5) {10};
    
    \node[scale=1.5] () at (-0.5, -1.5) {1:3};
    \node[scale=1.5] () at (-0.5, -4.5) {4:6};
    \node[scale=1.5] () at (-0.5, -7.5) {7:9};
\end{tikzpicture}
}
    }
    \hfill
    \subfloat[Structure of $\sum\limits_{(i,j) \in \mathds{C}}\widetilde{\mat{M}}_{k}^{i,j}$]{%
        \label{fig:matricStructure:second}%
        \def\colorPalleteRa{{32,62,74,57,28,46,45,56,21}}
\newcommand\ra{} 
\def\ra[#1](#2,#3)(#4,#5){%
    \foreach \x [count=\n from 0] in {0,1,2}
    {%
        \foreach \y [evaluate=\myMix using ({\colorPalleteRa[\n*3 + \y]}+20)]in {0, 1, 2} {%
            \draw [#1, draw=none, fill=matlabGreen!\myMix!black] (#2+#4*\x, #3-#5*\y) rectangle ++(#4,-#5);
        }
    }
}

\def\colorPalleteRaNeg{{32,62,74,57,28,46,45,56,21}}
\newcommand\ran{} 
\def\ran[#1](#2,#3)(#4,#5){%
    \foreach \x [count=\n from 0] in {0,1,2}
    {%
        \foreach \y [evaluate=\myMix using ({\colorPalleteRaNeg[\n*3 + \y]}+20)]in {0, 1, 2} {%
            \draw [#1, draw=none, fill=matlabRed!\myMix!black] (#2+#4*\x, #3-#5*\y) rectangle ++(#4,-#5);
        }
    }
}

\def\colorPalleteT{{90,60,77}}
\newcommand\tvec{} 
\def\tvec[#1](#2,#3)(#4,#5)(#6,#7){%
    \foreach \step [evaluate=\myMix using ({\colorPalleteT[\step]})]in {0,1,2} {%
        \draw [#1, draw=none, fill=#4!\myMix!black] (#2+#6*\step*#5, #3+#7*\step*#5) rectangle ++(#5,-#5);
    }
}

\resizebox{0.43\linewidth}{!}{%
\begin{tikzpicture}[baseline=(current bounding box.center)]
\useasboundingbox (-3,0.5) rectangle (11,-9.5);
    \node [scale=2] (rai) at (-4+2.7/2, -0) {$\mat{I} - \mat{R}_A$};
    \ran[](-4,-0.5)(0.9,0.9);

    \node [scale=2] (tb1) at (-4+2.7/2, -4) {$\vec{b}_{1} - \vec{b}_{2}$};
    \tvec[](-4,-4.5)(olive, 0.9)(1,0);
    \node [scale=2] (tb1) at (-4+2.7/2, -6) {$\vec{b}_{1} - \vec{b}_{3}$};
    \tvec[](-4,-6.5)(matlabPurple, 0.9)(1,0);
    \node [scale=2] (tb1) at (-4+2.7/2, -8) {$\vec{b}_{2} - \vec{b}_{3}$};
    \tvec[](-4,-8.5)(cyan, 0.9)(1,0);

    \ran[](0,0)(1,1);
    \ra[](3,0)(1,1);
    \ran[](0,-3)(1,1);
    \ra[](6,-3)(1,1);
    \ran[](3,-6)(1,1);
    \ra[](6,-6)(1,1);
    \tvec[](9,0)(olive, 1.0)(0,-1);
    \tvec[](9,-3)(matlabPurple, 1.0)(0,-1);
    \tvec[](9,-6)(cyan, 1.0)(0,-1);
    
    \draw[draw=black] (0,0) rectangle ++(10,-9);
    \draw[draw=black] (3,0) - ++(0,-9);
    \draw[draw=black] (6,0) - ++(0,-9);
    \draw[draw=black] (9,0) - ++(0,-9);
    \draw[draw=black] (0,-3) - ++(10,0);
    \draw[draw=black] (0,-6) - ++(10,0);

    \node[scale=1.5] () at (1.5, 0.5) {1:3};
    \node[scale=1.5] () at (4.5, 0.5) {4:6};
    \node[scale=1.5] () at (7.5, 0.5) {7:9};
    \node[scale=1.5] () at (9.5, 0.5) {10};
    
    \node[scale=1.5] () at (10.5, -1.5) {1:3};
    \node[scale=1.5] () at (10.5, -4.5) {4:6};
    \node[scale=1.5] () at (10.5, -7.5) {7:9};

\end{tikzpicture}
}
    }
    \caption{
        Inspired by~\cite{Briales2017ConvexDuality}, the structures of the matrices used to construct the cost functions with three antennas are illustrated.
        Therefore, the sum of matrices is considered at a singe time step (see~\eqref{eq:bigM}).
        Fig.~\ref{fig:matricStructure:first} relates to the matrix $\mat{M}^i$ of Eq.~\eqref{eq:Mkts}, and Fig.~\ref{fig:matricStructure:second} relates to the matrix $\widetilde{\mat{M}}_{k}^{i,j}$ of Eq.~\eqref{eq:Mcross}.
    }
    \label{fig:matricStructure}
\end{figure*}
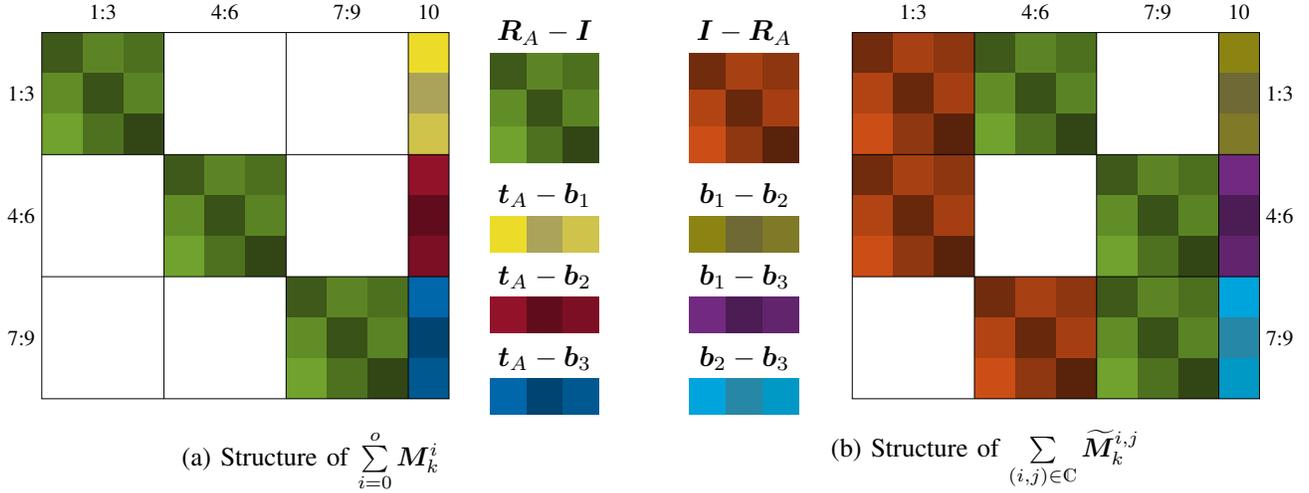
The derivation of the last section can be extended to allow for multiple antennas to be calibrated at once.
For this, the matrix $\mat{M}$ is inflated, similar to~\cite{horn2023extrinsic}.
Here, only desired antenna positions are added to the objective variable; a single homogenization variable $\mu$ is sufficient. 
Adjusting the derivation, given $o$ antennas, the matrix $\mat{M}_{k}^{i} \in \mathbb{R}^{3o \times n}$ at step $k$ for antenna $i$, the objective variable $\vec{z} \in \mathbb{R}^{n}$, and $n = 3o + 1$ is defined by
\begin{subequations}
\begin{align}
    \label{eq:Mkts}
    \mat{M}_{k}^{i} &:=
    \begin{bmatrix}
        \mat{E}_{i} \otimes (\mat{R}_{A_k} - \mat{I}) &
        \vec{e}_{i} \otimes (\vec{t}_{A_k} - \vec{b}_{i,k}) 
    \end{bmatrix} \pspace , \\
    \label{eq:z}
    \vec{z} &:= 
    \begin{bmatrix}
        \vec{x}_{0}\trans & \dots & \vec{x}_{o}\trans & 
        \mu
    \end{bmatrix}\trans \pspace ,
\end{align}
\end{subequations}
with $\mat{E}_{i} := \vec{e}_i \vec{e}_i\trans$, the canonical basis vectors $\vec{e}_{i} \in \mathds{R}^{3o}$, and the Kronecker product $\otimes$.
Again, the quadratic cost function $J: \mathbb{R}^{n} \rightarrow \mathbb{R} \pspace ; \pspace J(\vec{z}) := \vec{z}\trans \mat{Q} \vec{z}$ is constructed with $\mat{Q} \in \mathbb{R}^{n \times n}$ defined by
\begin{align}
    \mat{Q} := \sum_{(k, i) \in \mathds{D}} {\mat{M}_{k}^{i}}\trans \mat{M}_{k}^{i} \pspace ,
\end{align}
with $\mathds{D}$ containing all tuples of steps $k$ and antennas $i$.
Now, the solution to the optimization problem~\eqref{eq:primalProb} using the multi-antenna cost function without any quadratic constraints ($m_q = 0$) yields the result of a GNSS multi-antenna lever arm calibration.
Since the cost function does not contain any motion correlation between different antennas, each antenna calibration is considered separately during optimization.
Thus, calibrating multiple antennas simultaneously would not yet increase precision.
Therefore, in the multi-antenna case, an additional cycle can extend the objective function.

Given the scenario in Fig.~\ref{fig:gala} with two antennas, the following motion cycle does not directly take the IMU motion into account.
Instead, due to the rigid coupling, it is assumed that an antenna always rotates equally to the IMU sensor; hence, the translation measurements $\vec{b}_{i}$ of antennas $1$ and $2$ are considered as complete 6D transformations $B_{i}$.
Now, the transformation cycle is defined by
\begin{align}
\label{eq:trafoCycleAnt}
    (X_{2}\inv \circ X_{1} \circ B_{1})(\vec{0}) = (B_{2} \circ X_{2}\inv \circ X_{1})(\vec{0}) \pspace ,
\end{align}
where the position of antenna $2$ after motion is projected into the frame of antenna $1$ before motion using two different paths.
From~\eqref{eq:trafoCycleAnt}, the general case for antennas $i$ and $j$ is further derived using HMs to represent the transformations. 
With
\begin{align}
\label{eq:trafoCycleAntGen}
    X_{j}^{-1} \circ X_{i} = 
    \begin{bmatrix}
        \mat{I} & \vec{x}_{i} - \vec{x}_{j}\\
        \vec{0} & 1
    \end{bmatrix} \pspace ,
\end{align}
Eq.~\eqref{eq:trafoCycleAntGen} becomes
\begin{subequations}
\begin{alignat}{2}
    \begin{bmatrix}
        \mat{I} & \vec{x}_{i} - \vec{x}_{j}\\
        \vec{0} & 1
    \end{bmatrix} 
    \begin{bmatrix}
        \vec{b}_{i} \\
        1
    \end{bmatrix} 
    &=
    \begin{bmatrix}
        \mat{R}_A & \vec{b}_{j}\\
        \vec{0} & 1
    \end{bmatrix} 
    \begin{bmatrix}
        \vec{x}_{i} - \vec{x}_{j}\\
        1
    \end{bmatrix} \\[2pt]
    \label{eq:crossRaw}
    \vec{b}_{i} + \vec{x}_{i} - \vec{x}_{j} &= \mat{R}_A \vec{x}_{i} - \mat{R}_A \vec{x}_{j} + \vec{b}_{j} \pspace .
\end{alignat}
\end{subequations}
Rearranging Eq.~\eqref{eq:crossRaw} using $\vec{z}$ from Eq.~\eqref{eq:z} allows the integration in the previously defined optimization by extending the matrix $\mat{M}_{k}^{i}$ of Eq.~\eqref{eq:Mkts}.
The rearranged equation, with $\widetilde{\mat{M}}_{k}^{i,j} \in \mathbb{R}^{3l \times n}$, is
\begin{subequations}
\begin{gather}
    \widetilde{\mat{M}}_{k}^{i,j} \pspace \vec{z} = \vec{0} \quad \text{given}\\
    \label{eq:Mcross}
    \widetilde{\mat{M}}_{k}^{i,j} =
    \begin{bmatrix}
        \widehat{\mat{R}}_{A_k}^{p,j} - \widehat{\mat{R}}_{A_k}^{p,i} & \vec{e}_p \otimes (\vec{b}_{i,k} - \vec{b}_{j,k})
    \end{bmatrix}
\end{gather}
\end{subequations}
with the abbreviation $\widehat{\mat{R}}_{A_k}^{p,i} = \mat{E}_{p,i} \otimes (\mat{R}_{A_k} - \mat{I})$, the matrix $\mat{E}_{p,i} := \vec{e}_p \vec{e}_i\trans$, the number of possible combinations $l = \frac{o (o - 1)}{2}$, and a unique identifier $p \in [0,l]$ for each combination.
All matrices $\widetilde{\mat{M}}_{k}^{i,j}$ and $\mat{M}_{k}^{i}$ can now be combined to $\mat{M}_{k} \in \mathbb{R}^{(3o + l) \times n}$ denoted by
\begin{align}
\label{eq:bigM}
    \mat{M}_k = 
    \begin{bmatrix}
        \sum\limits_{i=0}^{o} \mat{M}_{k}^{i} \\
        \sum\limits_{(i,j) \in \mathds{C}}\widetilde{\mat{M}}_{k}^{i,j}
    \end{bmatrix} \pspace ,
\end{align}
where the set $\mathds{C}$ holds all combinations of antennas.
The upper and lower part structure of the matrix $\mat{M}_k$ is illustrated in Fig.~\ref{fig:matricStructure}.
Last, the matrix $\mat{Q} \in \mathbb{R}^{n \times n}$ of the extended cost function is defined by
\begin{align}
    \mat{Q} := \sum_{k} {\mat{M}_{k}}\trans \mat{M}_{k} \pspace .
\end{align}
As shown in the evaluation in Section~\ref{sec:exp}, the extended cost function acts as a regularization term, which helps the optimizer in the multi-antenna case when more constraints are involved.
Further, it helps to reduce the relative position error between antennas.

\subsection{Integration of Prior Knowledge}
The construction of cost functions and the respective optimization problem of the previous sections are solely based on sensor motion measurements.
However, additional knowledge about the robotic system to be calibrated is occasionally available.
Exemplarily, the authors of~\cite{stovner2019gnss} state that the distance between the IMU sensor and the GNSS antenna is easily obtainable; thus, it seems desirable to integrate such knowledge into the calibration procedure.
In the proposed calibration, additional prior knowledge can be incorporated by adding respective constraints to the optimization problem.
For the example above, the constraint must ensure a certain length of the calibration output.
Given the antennas' lever arm lengths $s_i \in \mathbb{R}$, and $\mat{P}_i \in \mathbb{R}^{n \times n} $ the respective constraints are defined by
\begin{subequations}
\begin{gather}
\label{eq:normConst}
    g_i(\vec{z}) := \vec{z}\trans \mat{P}_{i} \vec{z} \pspace , \quad \text{with }
    \mat{P}_{i}  := \operatorname{diag}(
        \mat{E}_i \otimes \mat{I}_{3}, -s_i^2   
    ) \\
    \Rightarrow g_i(\vec{z}) = 0 \Leftrightarrow ||x_i||_2 = s_i^2 \pspace .
\end{gather}
\end{subequations}
Similarly, knowledge about single entries can be taken into account.
For this, all entries but the one of interest in the eye matrix in~\eqref{eq:normConst} are set to zero.
It shall be mentioned that these constraints can only consider differences in a quadratic manner, meaning that whenever a single entry is considered, the optimization might obtain a negative value, which must be taken into account afterward.

\subsection{Motion Requirements}
\label{sec:motionRequirements}
Similar to other hand-eye calibration methods, there are motion requirements the calibration data must fulfill to ensure a unique and valid solution.
The authors of~\cite{tsai1988realtime} determine that motion with at least two non-colinear rotation axes is necessary to ensure the uniqueness of a solution for the hand-eye problem.
However, as a GNSS sensor does not measure rotations, these requirements change for the antenna lever arm problem.
Given a rotation along a specific axis, the position offset within a surface with a parallel normal axis can be determined by a single measurement in the hand-eye case; instead, only the radius of this offset can be obtained in the antenna lever arm case.
This indicates that additional motion requirements must be met.

Looking at Eq.~\eqref{eq:singleCycle}, the matrix $\mat{M}$ must have a full rank of $3$ to ensure a unique solution.
This leads to the exact motion requirement with at least two non-colinear rotation axes of the hand-eye problem.
However, to allow the derivation in 3D space, the orientation of the GNSS frame must equal the IMU frame orientation (see. Section~\ref{sec:problemFormulation}).
This alignment is often assumed to be known before calibration~\cite{hong2005observ, stovner2019gnss}.
In general, however, this alignment must be calibrated beforehand.
IMU sensors used in autonomous vehicles often align to the horizon without moving using a magnetometer and an accelerometer.
The only missing alignment is the yaw angle.
If available with the IMU sensor, this can be measured by a compass sensor~\cite{zhang2005navigation}.
Otherwise, a single translational motion without rotation or motion with a rotation around any axis but the \textit{up}-vector is sufficient to align both sensors in 3D space.

In summary, the antenna lever arm calibration proposed in this work requires motion with at least two non-colinear rotation axes and an aligned sensor set.

\subsection{Planar Motion}
\label{sec:planarMotion}
A common problem with motion-based calibration for autonomous vehicles is the restricted motion of the vehicles.
Acquiring motion data with two non-colinear rotation axes is difficult~\cite{horn2021online}.
As a result, the z-component of the calibration cannot be estimated accurately.
To enable a motion-based hand-eye calibration nonetheless, we proposed an extension to the hand-eye problem in previous works~\cite{horn2021online, horn2023extrinsic}.
Similarly, the antenna lever arm calibration extension is described next.

Additional knowledge must be available to allow a calibration based on planar motion.
More specifically, the lever arm length or its z-component must be known for each antenna.
In both cases, a norm constraint described in Eq.~\eqref{eq:normConst} is used to integrate the knowledge into the calibration process.
Like in~\cite{horn2023extrinsic}, using a norm constraint leads to two possible solutions along the \textit{up}-vector.
Thus, after optimization, the result must be reviewed.
Assuming that the antenna is always located above the IMU sensor, the correct calibration can be chosen by examining the z component.

\section{Experiments}
\label{sec:exp}
This section describes the evaluation process first, and respective results are presented and discussed afterward.
The different structure of the proposed approach and, thus, other input data do not allow a direct, fair comparison to state-of-the-art approaches. 
Instead, simulated data is used to compare results, considering different noise levels and motion restrictions. 
Then, the general progression of errors is compared to~\cite{stovner2019gnss} considering their noise assumptions.

The translation errors $\varepsilon_t$ between the ground truth calibration $\vec{t}_T$ and the predicted calibration $\widehat{\vec{t}}$ are calculated as
\begin{align}
    \varepsilon_t = || \widehat{\vec{t}} - \vec{t}_T ||_2 \pspace .
\end{align}
If not differently stated, all results are averaged over $10\,000$ simulation runs.
All evaluations were run on a computer running Ubuntu 22.04 containing an ADM\,Ryzen\textsuperscript{TM}\,Threadripper\textsuperscript{TM}\,3970X CPU and 128GB of DDR4 RAM. 
Each calibration is executed single-threaded; therefore, run times are not sped up using more CPU threads.

\subsection{Data Simulation}
\begin{figure}[t]
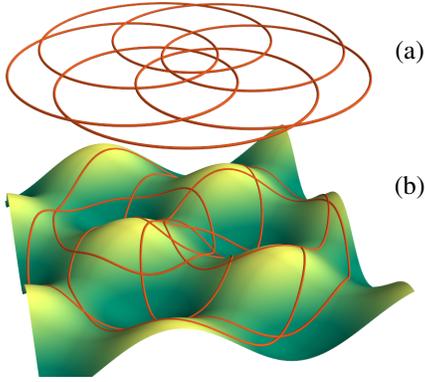

    \centering
    \vspace*{5mm}
    \include{img/surface_path}
    \vspace*{-5mm}
    \caption{
        A simulated path example:
        The 2D path (a), a sinusoid mixture in this example, is projected onto a rough surface, resulting in a 3D path (b).
    }
    \label{fig:simulationPath}
\end{figure}
Data is simulated similarly to our previous work~\cite{horn2021online}.
For this, a 2D path is projected onto a 3D surface, illustrated in Fig.~\ref{fig:simulationPath}.
Given the gradients of both path and surface, the orientations can be calculated for each point leading to a 6D path of poses.
Motion data with different rotation axes, called \method{Hilly} in the following, can be generated using the hilly surface depicted in Fig.~\ref{fig:simulationPath}.
A flat surface is also used to generate planar motion data, called \method{Flat}.
Antenna motion for up to three antennas is generated, given a ground-truth transformation for each antenna with an antenna arm length of $\SI{1}{\metre}$.
Further, unbiased Gaussian noise can be added to the poses.
Its standard derivation, called noise level, is set with absolute or relative values.
For example, if the average motion along the path travels $\SI{1}{\metre}$ and $\SI{0.5}{\radian}$, a relative noise level of $\SI{10}{\percent}$ leads to Gaussian noise with $\sigma_{\text{trans}} = \SI{0.1}{\metre},\, \sigma_{\text{rot}} = \SI{0.05}{\radian}$.

\subsection{Single Antenna Calibration}

\begin{figure}
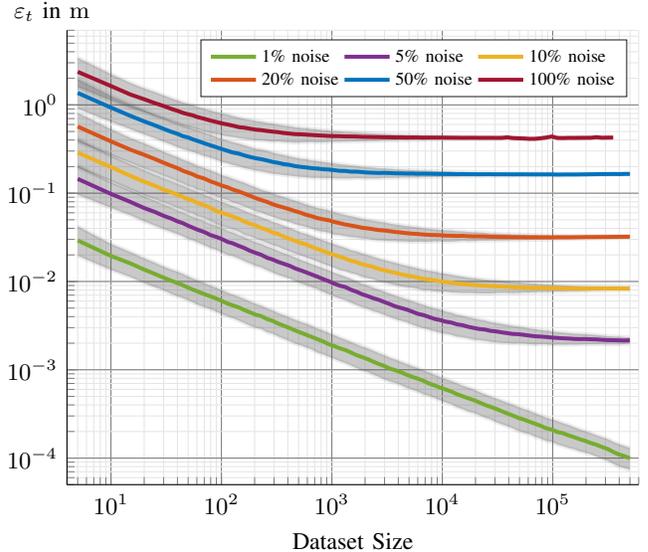

	\centering
    \vspace*{1mm}
%
%
%
\begin{tikzpicture}[baseline=(current bounding box.center)]

\begin{axis}[%
width=0.88\linewidth,
height = 0.7\linewidth,
scale only axis,
xmode=log,
xmin=4,
xmax=600000,
xminorticks=true,
xlabel style={font=\color{white!15!black}, font=\small},
xlabel={Dataset Size},
ymode=log,
ymin=0.00005,
ymax=7,
yminorticks=true,
ylabel style={font=\color{white!15!black}, font=\small,at={(current axis.north west)},above=2mm, rotate=-90},
ylabel={$\varepsilon_t$ in \si{\metre}},
axis background/.style={fill=white},
axis x line*=bottom,
axis y line*=left,
xmajorgrids,
xminorgrids,
ymajorgrids,
yminorgrids,
major grid style={black!50},
minor grid style={black!10},
legend columns = 3,
ticklabel style = {font=\small},
legend style={font = \small, legend cell align=left, align=left, draw=white!15!black,nodes={scale=0.75, transform shape}}
]

\colorlet{1Color}{matlabGreen}
\colorlet{5Color}{matlabPurple}
\colorlet{10Color}{matlabYellow}
\colorlet{20Color}{matlabRed}
\colorlet{50Color}{matlabBlue}
\colorlet{100Color}{matlabDarkRed}

\input{img/raw/calibration_mean_noise/1_tikz_points.txt}
\input{img/raw/calibration_mean_noise/5_tikz_points.txt}
\input{img/raw/calibration_mean_noise/10_tikz_points.txt}
\input{img/raw/calibration_mean_noise/20_tikz_points.txt}
\input{img/raw/calibration_mean_noise/50_tikz_points.txt}
\input{img/raw/calibration_mean_noise/100_tikz_points.txt}

\end{axis}
\end{tikzpicture}
	\caption{
        Translation errors of calibrations based on simulated data are plotted over the number of points used for the optimization.
        A line denotes the $50$th quantile over $10000$ simulation runs for the respective noise level.
        The gray area behind each line denotes the respective $25$th to $75$th quantile.
        All curves show a converging behavior dependent on the respective noise level.
    }
	\label{fig:simulatedResults}
\end{figure}

First, we used simulated \method{Hilly} data to demonstrate the general behavior of our single-antenna approach for various noise levels and data set sizes. 
The antenna arm length is $\SI{1}{\metre}$.
Results are illustrated in Fig.~\ref{fig:simulatedResults}.

Each colored line represents the error progression of a certain noise level over the data set size.
The general progression is similar to~\cite{horn2021online, stovner2019gnss} but smoother due to more simulation runs.
Each noise level has a lower bound to which the error converges.
With a higher noise level, this bound is higher but is reached with fewer data points. 
Thus, the calibration error can not constantly be improved by adding more data to the optimization; instead, the best possible error depends on the noise level.
The overall error progression in Fig.~\ref{fig:simulatedResults} shows that our approach can robustly obtain a feasible lever arm calibration, even under the influence of strong noise.
In all cases, given enough data points, the ratio between the error and the antenna arm length is lower than the respective noise level.

\subsection{Multi-Antenna Calibration}
We evaluated the multi-antenna calibration using \method{Hilly} motion with $\SI{10}{\percent}$ noise level and a data set size of $5\,000$.
For this setting, the errors of the single antenna case are not yet converging, which allows us to demonstrate the influence of different settings in the multi-antenna case.
Fig.~\ref{fig:multiAntennaResults:equal} shows the GNSS multi-antenna lever arm calibration results.
The calibration was performed with four different settings for each number of antennas.
The first two settings (\rom 1, \rom 2) use the optimization described in Section~\ref{sec:multiAntennacalibration}; only in the second the regularization terms are added. 
Then, prior knowledge of the antenna lever arm length is incorporated into the calibration for the other two settings (\rom 3, \rom 4); again, the regularization terms are only added to the latter (\rom 4).

The results show that using additional knowledge improves the calibration.
Without the regularization term, the mean translation error of the calibration is relatively constant with an increasing amount of antennas.
Considering regularization terms does not improve the result.
The overall improvement using prior knowledge is $\SI{27}{\percent}$, $\SI{42}{\percent}$, and $\SI{60}{\percent}$ for $1$, $2$, and $3$ antennas respectively.

The noise level was altered for the next experiment to demonstrate the advantages of the regularization term.
Instead of setting an equal noise level for the IMU and GNSS sensors, the rotational noise level of the IMU was divided by six, and the translational noise level was sixfold.
This change is motivated by the survey~\cite{samatas2022inertial} in which several IMU sensors are compared, and most show a higher translational than rotational noise.
The results are shown in Fig.~\ref{fig:multiAntennaResults:unequal}
Here, with more antennas, the regularization term significantly improves the calibration.
The overall improvement using the regularization and prior knowledge is $\SI{21}{\percent}$, $\SI{43}{\percent}$, and $\SI{67}{\percent}$ for $1$, $2$, and $3$ antennas respectively.

In our experiments, the relative error between antenna positions was smaller when using the regularization and a known antenna arm length.
However, the magnitude of this effect was strongly dependent on the antenna position and the type of motion.
Thus, for a reliable statement, the impact needs to be investigated in more detail regarding the antenna distribution, and it is up for future development.
\begin{figure}
    \vspace*{3mm}
    \input{img/raw/multi_antenna/boxplot_tikz_data}
    \centering
    \captionsetup[subfloat]{labelfont=normalsize,textfont=normalsize}
    \subfloat[Equal Noise Level]{%
        \label{fig:multiAntennaResults:equal}%
%
%
%
\begin{tikzpicture}[baseline=(current bounding box.center)]

\input{img/boxplot}

\begin{axis} [
    width=\linewidth,
    height = 0.5\linewidth,
    scaled y ticks = false,
    box plot width=1mm,
    xlabel={\small\# Antennas},
    xlabel style={yshift=1mm},
    xtick={0,2,5,6,7,8,10,11,12,13}, 
    xticklabels={%
        {\rom 1},{\rom 3},%
        {\rom 1},{\rom 2},{\rom 3},{\rom 4},%
        {\rom 1},{\rom 2},{\rom 3},{\rom 4}%
    },
    xticklabel style={font=\footnotesize},
    extra x ticks={1.5, 6.5, 11.5},
    extra x tick labels={{1},{2},{3}},
    extra x tick style={%
        tick label style={%
            yshift=-4mm, %
        }, %
        xticklabel style={font=\normalsize},%
        tick style={%
            draw=none%
    }},
    ylabel={\small$\varepsilon_t$ in \si{\centi\meter}},
    ymin=-0.0025,
    ymax=0.025,
    ytick={0.0, 0.01, 0.02}, 
    yticklabels={{0},{1}, {2}},
    ymajorgrids,
    yminorgrids,
]
\boxplot {backGray}{matlabGreen} {multi_antenna_equal_distr.dat}

\end{axis}
\end{tikzpicture}%
    }\\
    \subfloat[Unequal Noise Level]{%
        \label{fig:multiAntennaResults:unequal}%
%
%
%
\begin{tikzpicture}[baseline=(current bounding box.center)]

\input{img/boxplot}

\begin{axis} [
    width=\linewidth,
    height = 0.5\linewidth,
    scaled y ticks = false,
    box plot width=1mm,
    xlabel={\small \# Antennas},
    xlabel style={yshift=1mm},
    xtick={0,2,5,6,7,8,10,11,12,13}, 
    xticklabels={%
        {\rom 1},{\rom 3},%
        {\rom 1},{\rom 2},{\rom 3},{\rom 4},%
        {\rom 1},{\rom 2},{\rom 3},{\rom 4}%
    },
    xticklabel style={font=\footnotesize},
    extra x ticks={1.5, 6.5, 11.5},
    extra x tick labels={{1},{2},{3}},
    extra x tick style={%
        tick label style={%
            yshift=-4mm, %
        }, %
        xticklabel style={font=\normalsize},%
        tick style={%
            draw=none%
    }},
    ylabel={\small$\varepsilon_t$ in \si{\centi\meter}},
    ymin=-0.005,
    ymax=0.13,
    ytick={0.0, 0.025, 0.05, 0.075, 0.1, 0.125},
    yticklabels={{0},{2.5},{5},{7.5},{10},{12.5}},
    ymajorgrids,
    yminorgrids,
]
\boxplot {backGray}{matlabGreen} {multi_antenna_unequal_distr.dat}

\end{axis}
\end{tikzpicture}%
    }
	\caption{%
        Calibration errors are shown for different calibration settings with varying numbers of antennas.
        Calibration settings (\rom 1) and (\rom 2) use no prior knowledge, and (\rom 3) and (\rom 4) incorporate the knowledge about the lever arm length.
        Calibration settings (\rom 2) and (\rom 4) use the regularization term.
        In~\protect\subref{fig:multiAntennaResults:equal}, the noise levels are set equally for IMU and GNSS sensors, whereas in~\protect\subref{fig:multiAntennaResults:unequal}, the noise levels are set differently.%
    }
	\label{fig:multiAntennaResults}
\end{figure}

\subsection{Planar Motion}
To show the advantages of the planar motion extension, we evaluated our approach on both \method{Hilly} and \method{Flat} motion data.
Additionally, we used the KITTI odometry dataset~\cite{geiger2012kitti} and extracted the motion; the respective path is called \method{Kitti} in the following. 
Using the sequences $04-10$ from \method{Kitti}, a total of approx. $12\,000$ motion steps are available.
This allows us to test the approach with real-world IMU sensor motion captured from an autonomous vehicle.
Antenna motion is simulated with $\SI{10}{\percent}$ noise level.
We used a dataset size of $10\,000$ to evaluate the planar motion approach, allowing us to draw slightly different paths from \method{Kitti} for each run.
The results are presented in Table~\ref{tab:planarMotion} together with those using the \method{Hilly} and \method{Flat} data.
In addition to the four settings before, we added a fifth setting (\rom 5), which knows the height of the sensor and integrates this knowledge into the calibration.
Further, for better readability and due to their similar performance in the previous section, settings (\rom 2) and (\rom 3) are omitted.
To emphasize the necessity of the planar motion extension, the calibration results using the path generated from a hilly surface are shown in the first line of Table~\ref{tab:planarMotion} as a reference.
\begin{table}[t]
    \vspace*{3mm}
    \caption{
        The planar motion calibration results (mean translation error $\varepsilon_t$ in $\si{\centi\metre}$) for various antennas~(1-3) and different calibration settings (\rom 1, \rom 4, \rom 5).
    }
    \label{tab:planarMotion}
    \begin{center}
    \begin{threeparttable}
        \setlength{\tabcolsep}{0.5em}
        \begin{tabular}{lO{2.1}O{2.1}O{2.1}|O{2.1}O{2.1}O{2.1}|O{2.1}O{2.1}O{2.1}}

\toprule
&
\multicolumn{3}{c}{1} &
\multicolumn{3}{c}{2} &
\multicolumn{3}{c}{3} 
\\

\textbf{Path} &
\rom 1 & \rom 4 & \rom 5 &
\rom 1 & \rom 4 & \rom 5 &
\rom 1 & \rom 4 & \rom 5 
\\

\midrule

\method{hilly} & 1.2 & 0.6 & 0.4 & 1.2 & 0.6 & 0.4 & 1.2 & 0.5 & 0.2  \\
\method{flat} & 93.0 & 1.6 & 0.9 & 93.0 & 1.1 & 0.6 & 95.0 & 0.9 & 0.1  \\
\method{kitti} & 11.9 & 11.3 & 10.3 & 11.9 & 5.5 & 4.1 & 12.0 & 5.8 & 1.5  \\

\bottomrule
    
\end{tabular}

    \end{threeparttable}
    \end{center}
\end{table}

Generally, results with settings (\rom 4) and (\rom 5) become better with an increasing number of antennas.
Comparing the results of the first setting (\rom 1) on different paths confirms the claim of Section~\ref{sec:planarMotion} that an accurate calibration based on planar motion is impossible without further knowledge.
With enough rotation around the up-vector, the calibration on flat surface motion becomes feasible using the fourth setting (\rom 4) with the knowledge of the lever arm length.
It is further improved with the fifth setting (\rom 5).
This shows that our proposed approach can overcome the issue of insufficient rotation axes in planar motion.
However, settings (\rom 4, \rom 5) based on \method{Kitti} still lack some precision compared to the other methods.
In contrast, setting (\rom 1) performs better on \method{Kitti} than on \method{Flat}.
Compared to \method{Flat}, which is strictly limited to a single rotation axis, the motion of \method{Kitti} includes more rotation axes, yet with small amplitudes, which explains the better performance with setting (\rom 1).
However, the reduced performance increase of \method{Kitti} with more antennas compared to \method{Flat} is most likely due to primarily translational motion of in \method{Kitti}.
An exemplary path of the \method{Kitti} motion showing this phenomenon is illustrated in~\cite{horn2021online}.
Nevertheless, our proposed extension enables a feasible solution in this case. 
This indicates that integrating knowledge compensates for not only the lack of rotation axes but also motion restrictions in general.

In this section, the exact lever arm length was known and integrated as prior knowledge into the calibration process.
However, in real-world experiments, this information is subject to errors.
Additional experiments have been conducted, but due to limited space available, they could not be included here.
Nevertheless, it shall be mentioned that in these experiments the increase in translational calibration error was roughly equivalent to the error of the lever arm length.

Summarizing the results on planar motion calibration, the flat motion restriction of autonomous vehicles can be overcome using our proposed extension.
The results indicate that a viable real-world calibration requires enough rotational motion around a single axis.
However, our method finds a feasible solution even with less rotational motion in the multi-antenna case.
Adding a suitable sample weighting similar to~\cite{horn2023user} in the future could further improve autonomous vehicle calibration based on primarily translational motion.

\subsection{Run Times}
\begin{figure}
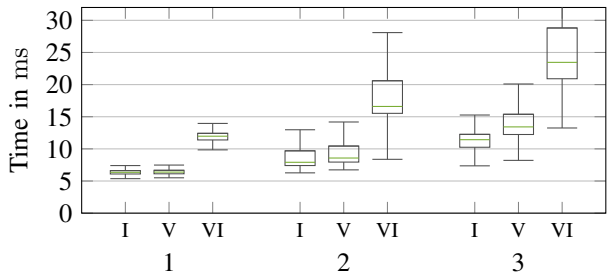

    \vspace*{3mm}
	\centering
%
%
%
\begin{tikzpicture}[baseline=(current bounding box.center)]

\input{img/boxplot}
\input{img/raw/timings/timings_boxplot_tikz_data}

\begin{axis} [
    width=0.99\linewidth,
    height = 0.5\linewidth,
    scaled y ticks = false,
    box plot width=2mm,
    xtick={0,1,2,4,5,6,8,9,10}, 
    xticklabels={%
        {\rom 1},{\rom 5},{\rom 6},%
        {\rom 1},{\rom 5},{\rom 6},%
        {\rom 1},{\rom 5},{\rom 6},%
    },
    xticklabel style={font=\footnotesize},
    extra x ticks={1, 5, 9},
    extra x tick labels={{1},{2},{3}},
    extra x tick style={%
        tick label style={%
            yshift=-4mm, %
        }, %
        xticklabel style={font=\normalsize},%
        tick style={%
            draw=none%
    }},
    ylabel={Time in \si{\milli\second}},
    ymin=0,
    ymax=0.032,
    ytick={0.0, 0.005, 0.01, 0.015, 0.02, 0.025, 0.03},
    yticklabels={{0},{5},{10},{15},{20},{25},{30}},    
    ymajorgrids,
    yminorgrids,
]
\boxplot {backGray}{matlabGreen} {timings.dat}

\end{axis}
\end{tikzpicture}
	\caption{
        Calibration run times for different numbers of antennas~(1-3) and calibration settings (\rom 1, \rom 4, \rom 5) are shown. 
        For illustration reasons, the upper whisker of the last box is outside the plotting area; it is located at $\SI{40}{\micro\second}$.
    }
	\label{fig:timings}
\end{figure}

Last, we evaluated the calibration run time.
In general, the run time varies with different settings.
For simplicity reasons, using the \method{Hilly} data, only the results for specific settings (\rom 1, \rom 4, \rom 5) are shown in Fig.~\ref{fig:timings}, since they represent the simplest and most complex configurations, respectively.
For the run-time evaluation, each calibration was run $100\,000$ times.
To limit the influence of UI tasks, the system nice level of the evaluation process was set to $-19$. 

With an increasing number of antennas, the calibration run-time gets more efficient relative to the amount of antennas calibrated.
The cost of incorporating the knowledge about the antenna lever arm is comparatively small.
In contrast, taking the height of an antenna into account at the same time is costly; run times increase by approximately factor two.
In general, all settings with up to three antennas allow high update rates from $\SI{40}{\hertz}$ up to more than $\SI{100}{\hertz}$, which exceeds usual sensor update rates in autonomous vehicles.
Additionally, the illustrated run times describe the calibration time from scratch; an online scheme usually allows even faster update rates due to only small changes between cycles.
Thus, our approach with all extensions is suitable for an online antenna lever arm calibration in autonomous vehicles.
\section{Conclusion}
In this work, we have derived a QCQP formulation, including a regularization for the multi-antenna lever arm calibration.
We analyzed the problem structure and stated motion requirements to ensure a valid and unique solution.
Overcoming motion restrictions of autonomous vehicles, we extended our approach to enable calibration based on planar motion.
With an extensive evaluation, we demonstrated that our approach obtains precise calibration results, and the fast run time allows us to use our approach in an online scheme in future works.

\section*{Acknowledgment}
The authors want to thank their former colleague, Markus Horn, for the fruitful discussions, which helped to improve this work.

\bibliographystyle{IEEEtran}
\bibliography{IEEEabrv,references}

\end{document}